\documentclass[runningheads]{llncs}
\usepackage{array}
\usepackage{booktabs}
\usepackage{multirow}
\usepackage[table,xcdraw]{xcolor}
\usepackage{marvosym}
\usepackage{graphicx}
\usepackage{amsmath}
\usepackage[ruled]{algorithm2e}
\makeatletter
\def\thanks{\protected@xdef\@thanks{\@thanks\protect\footnotetext{\dag\ Equal contributions}}}
\makeatother

\begin{document}
\title{Shallow Attention Network for Polyp Segmentation}

% \author{Anonymous}
% \institute{Paper ID: 2335}

\author{
    Jun Wei\inst{1,2,\dag},
    Yiwen Hu\inst{1,2,5,\thanks{\dag}},
    Ruimao Zhang\inst{1,2},
    Zhen Li\inst{1,2,\textrm{\Letter}},\\
    S.Kevin Zhou\inst{1,3,4},
    Shuguang Cui\inst{1,2}
}

% $^3$ University of Science and Technology of China \\
\institute{
    $^1$ School of Science and Engineering, The Chinese University of Hong Kong(Shenzhen) \\
    $^2$ Shenzhen Research Institute of Big Data \\
    $^3$ School of Biomedical Engineering \& Suzhou Institute for Advanced Research, University of Science and Technology of China, Suzhou, China\\
    $^4$ Institute of Computing Technology, Chinese Academy of Sciences, Beijing, China\\
    $^5$ Institute of Urology, The Third Affiliated Hospital of Shenzhen University(Luohu Hospital Group), Shenzhen, China.\\
    \email{lizhen@cuhk.edu.cn}
}

\maketitle

\begin{abstract}
Accurate polyp segmentation is of great importance for colorectal cancer diagnosis. However, even with a powerful deep neural network, there still exists three big challenges that impede the development of polyp segmentation. (i) Samples collected under different conditions show inconsistent colors, causing the feature distribution gap and overfitting issue; (ii) Due to repeated feature downsampling, small polyps are easily degraded; (iii) Foreground and background pixels are imbalanced, leading to a biased training. 
To address the above issues, we propose the {\textbf{S}}hallow {\textbf{A}}ttention {\textbf{Net}}work ({\textbf{SANet}}) for polyp segmentation. Specifically, to eliminate the effects of color, we design the color exchange operation to decouple the image contents and colors, and force the model to focus more on the target shape and structure. 
Furthermore, to enhance the segmentation quality of small polyps, we propose the shallow attention module to filter out the background noise of shallow features. Thanks to the high resolution of shallow features, small polyps can be preserved correctly.
In addition, to ease the severe pixel imbalance for small polyps, we propose a probability correction strategy (PCS) during the inference phase. Note that even though PCS is not involved in the training phase, it can still work well on a biased model and consistently improve the segmentation performance. 
Quantitative and qualitative experimental results on five challenging benchmarks confirm that our proposed SANet outperforms previous state-of-the-art methods by a large margin and achieves a speed about \textbf{72FPS}.
\end{abstract}

\section{Introduction}
\label{introduction}
Colorectal Cancer (CRC) has become a serious threat for human health, causing the fourth highest cancer death rate worldwide~\cite{favoriti2016worldwide}. Polyps in the intestinal mucosa are considered the harbinger of CRC, which are easily transformed into malignant lesions~\cite{granados2017colorectal}. Therefore, early diagnosis and treatment of polyps are of great significance. Fortunately, with the assistance of computer technology, a lot of automatic polyp segmentation models~\cite{akbari2018polyp,brandao2017fully,fan2020pranet,fang2019selective,mamonov2014automated,ronneberger2015u,tajbakhsh2015automated,zhang2020adaptive,zhang2018road,zhou2018unet++} have been developed and achieved remarkable progress.

\begin{figure}[t]
  \centering
  \includegraphics[scale=0.35]{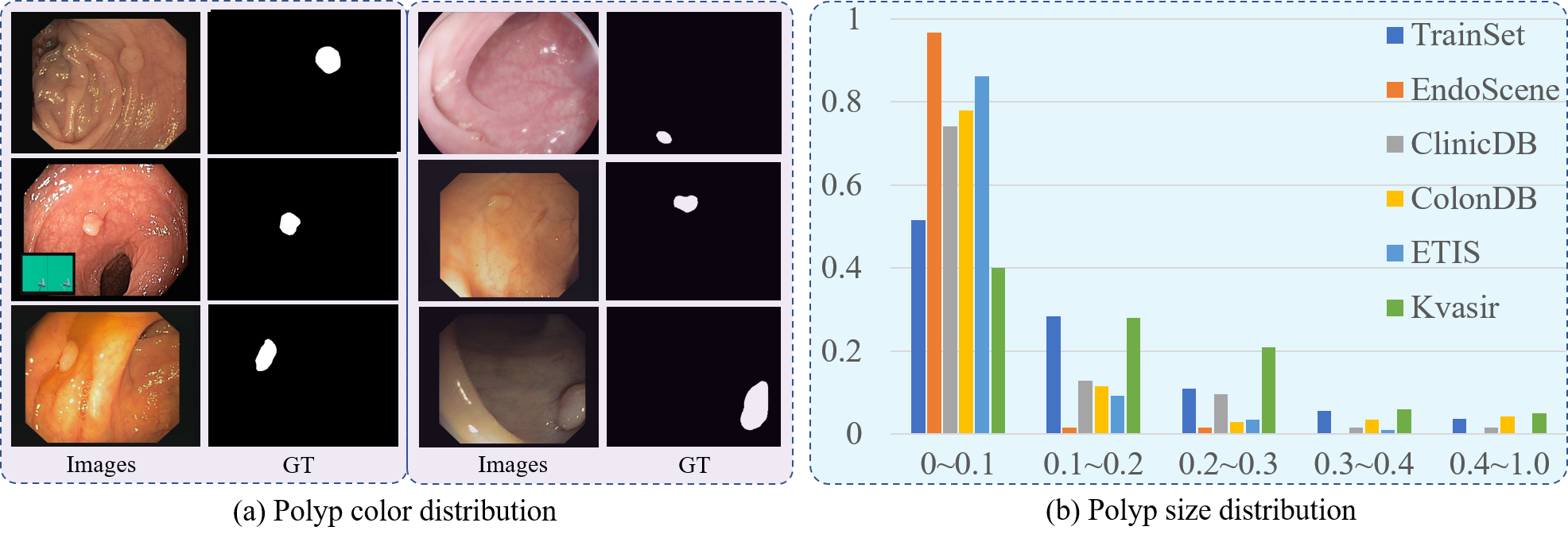}
  \caption{(a) Polyp samples with different colors. (b) The histogram of polyp size. The horizontal axis shows the proportion of the polyp area to the image area. The vertical axis shows the proportion of polyp samples of a specific size to the total samples. One training set and five testing sets are shown in different colors.}
  \label{fig:distribution}
\end{figure}

However, polyp segmentation has always been a challenging task due to the inconsistent color distribution and small size of the targets. As shown in Fig.~\ref{fig:distribution}(a), polyp samples collected under the same conditions usually correspond to the same color, while those collected under different conditions appear in different colors. Thus, a strong correlation between color and polyp segmentation is implicitly contained in the dataset, which is harmful to the model training and will cause the model to overfit the color.
Besides, most of polyp areas are very small. As shown in Fig.~\ref{fig:distribution}(b), the vast majority of polyps have an area which is less than 0.1 of the total image area. This brings two difficulties for existing segmentation models. First, small polyps are prone to get lost and hard to restore because of the repeated feature downsampling. Second, for images with small polyps, there exists a large imbalance between foreground and background pixels, leading to a biased model and a poor performance. 

To break the correlation between color and polyp segmentation, we propose the color exchange (CE) operation. Specifically, for each input image, we randomly select another one and transfer its color to the input image. By multiple CE operations, each input image could appear in different colors, thus the connection between color and polyp segmentation is reduced and the model will not overfit to the fake causality. 
Besides, small polyp segmentation relies more on shallow features. Because they have higher resolutions and contain richer details, compared with deep ones~\cite{wei2020f3net}. Unfortunately, shallow features are too noisy to be used directly. Thus, we propose the shallow attention module (SAM), which could help to remove the background noise, using clearer deep features.
Furthermore, to ease the imbalance between foreground and background, we propose the probability correction strategy (PCS) during inference, which can adjust the biased prediction to the correct one with simple post-processing. 

In summary, our contributions are four-folds: (1) We propose the color exhange operation to decouple the contents and colors, which reduces the overfitting issue; (2) We design the novel SANet to focus more on shallow features with richer details, improving the small polyp segmentation; (3) We introduce the probability correction strategy (PCS) for inference to balance the biased predictions, which can consistently improve model performance with negligible computation cost; (4) Extensive experiments demonstrate that the proposed model achieves state-of-the-art performace on five widely used public benchmarks.

\section{Related Work}
\begin{figure}[t]
    \centering
    \includegraphics[scale=0.335]{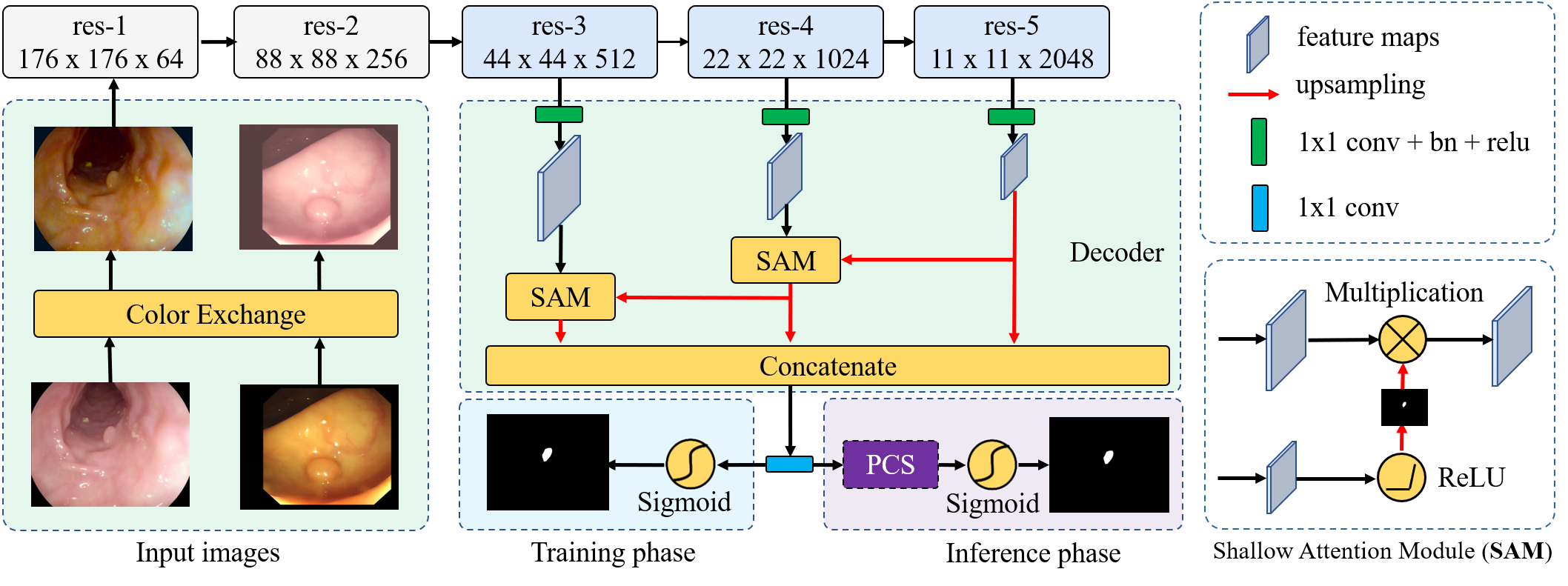}
    \caption{An overview of the proposed model.}
    \label{fig:framework}
  \end{figure}
Polyp segmentation has went through two periods of development. Earlier models mainly rely on hand-crafted features (e.g., color and texture)~\cite{mamonov2014automated,zhou2020review,tajbakhsh2015automated}, which can hardly capture the global context information and are not robust to complex scenarios. Recently, fully convolutional network (FCN~\cite{long2015fully}) has been applied to polyp segmentation and made great progress. For example, U-Net~\cite{ronneberger2015u} is a famous structure for medical image segmentation, which consists of a contracting path to capture context and an expanding path to restore the precise detail. SegNet~\cite{segnet} adopts a similar structure to U-Net, but utilizes the max pooling indices to restore features in the upsampling operation. U-Net++~\cite{zhou2018unet++} and ResUNet++~\cite{jha2019resunet++} further improve original U-Net by dense connection and better pretrained backbone, achieving promising segmentation performance.

Though the body part of polyps could be well handled, boundaries are ignored. To enhance the boundary segmentation, Psi-Net~\cite{murugesan2019psi} proposes to combine both body and boundary features in the segmentation model. Furthermore, SFA~\cite{fang2019selective} explicitly applies an area-boundary constraint to supervise the learning of both polyp regions and boundaries. PraNet~\cite{fan2020pranet} proposes the reverse attention to firstly locate the polyp areas and then refine object boundaries implicitly. Though the above models have made great progress, polyp segmentation still faces big challenges because of the limited data. Thus, we propose the SANet to further improve the polyp segmentation, as shown in the Fig.~\ref{fig:distribution}.

\section{Method}
Fig.~\ref{fig:framework} depicts the concrete architecture of the proposed SANet, where Res2Net~\cite{GaoCZZYT21} is adopted as the encoder backbone. According to the feature scale, Res2Net could be divided into five blocks, which have different receptive fields. Considering Wu {\it et al.}~\cite{CPD} have shown that low-level features bring too much computational cost with limited performance gains, thus we use the features of last three blocks $\{f_i | i \in (3,4,5) \}$ for the following experiments. 

\subsection{Color Exchange}
\begin{algorithm}[t]
    \caption{Color Exchange}
    \label{ColorExchange}
    \LinesNumbered
    \KwIn{Img1, Img2}
    \KwOut{Out1, Out2}
    transform Img1, Img2 from RGB space to LAB space, then get Lab1, Lab2\\
    calculate the channel mean and channel std of Lab1 and Lab2\\
    Lab1 = (Lab1-mean1)/std1*std2+mean2\\
    Lab2 = (Lab2-mean2)/std2*std1+mean1\\
    transform Lab1, Lab2 from LAB space to RGB space, then get Out1, Out2
\end{algorithm}
As show in Fig.~\ref{fig:distribution}(a), polyp samples collected under different conditions show very different color distributions. However, these colors are far less important than shapes and structures for polyp segmentation, which actually leads to the overfitting issue due to the limited dataset. To avoid this effect, we propose the color exchange (CE) operation to explicitly decouple the image content and color. Specifically, for each input image, we randomly pick another one from the dataset and transfer its color to the input image. Alg.~\ref{ColorExchange} shows the specific steps. We calculate the mean and standard deviation of the colors in the LAB space and exchange these statistics between images. As show in Fig.~\ref{fig:framework}, after color exchange, we could get the new input image with exactly the same content but a different color. Exchanging with different auxiliary images, the same input image could show a variety of colors but correspond to the same ground truth, so the model will focus more on the image contents and will not be affected by the color distribution, which could largely alleviate the influence of color distribution on model training. It is worth noting that color exchange is performed only during training. For inference, we directly use the original images, thus it will not bring any time overhead during testing.

\subsection{Shallow Attention Module}
As shown in Fig.~\ref{fig:distribution}(b), small polyps make up the majority of both training and testing datasets, facing the serious information loss during the repeated downsampling in CNNs. To avoid this limitation, shallow features $f_s$ deserve more attention due to their high resolutions. Namely, $f_s$ have clear object boundaries, which are important for the accurate polyp segmentation. However, due to the limitation of receptive field, these features are submerged by background noise and hard to be used directly. In contrast, deep features $f_d$ are coarse in boundaries but have clean background. Therefore, we propose the shallow attention module (SAM) to filter out background noise of $f_s$ with the assistance of $f_d$, as shown in Fig.~\ref{fig:framework}.
Specifically, SAM involves in both $f_s$ and $f_d$. Different from self-attention, SAM makes use of the complementarity between different features. In SAM, $f_d$ will be firstly upsampled into the same size with $f_s$ and then regarded as the attention maps for $f_s$. Finally, these maps will be multiplied with $f_s$ to help suppress the background noise. The whole process is shown in following equations:
\begin{align}
Att &= \sigma(Up(f_d)), \\
f_s &= Att \otimes f_s,
\end{align}
where $Up(\cdot)$ represents the upsampling operation. $\sigma(\cdot)$ is the ReLU function and $\otimes$ is the element-wise multiplication. After SAM, shallow features $f_s$ will become much cleaner and provide important cues for the segmentation of small polyps. Furthermore, SAM plays an important role in balancing features of different blocks. Instead of aggregating features from all levels with the same weight, SAM could dynamically assign weight to different features according to their contributions.

\subsection{Probability Correction Strategy}
Probability correction strategy (PCS) is an effective way to improve final predictions, especially for small polyps where foreground and background pixels are extremely unbalanced. This imbalance brings in difficulties for the model training and leads to a biased model. Namely, negative samples (background pixels) are dominant in the training process, which leads to the tendency of the model to give lower confidence to positive samples (foreground pixels). To enhance the predictions of the positive samples, we propose to explicitly correct the predicted probability through logit reweighting. Specifically, for each image, we extract the predicted features before the Sigmoid function as the target ({\it i.e.,} logit) to be corrected. We count the proportion ($rate^p$) of positive samples ({\it i.e.,} $logit>0$) in the image, which is relatively small. At the same time, due to the model bias, the logit of the positive sample also is small. Thus, we propose to normalize the logit of positive samples with its proportion, as shown in Eq.~\ref{eq:normal_pos}. 

\begin{equation}
logit^{p_{norm}}_{ij} = logit^p_{ij}/rate^p \label{eq:normal_pos}
\end{equation}
where $i,j$ represent the coordinate of the sample. Similarly, the logits of negative ({\it i.e.,} $logit<0$) samples could also be normalized, as shown in Eq.~\ref{eq:normal_neg}.

\begin{equation}
logit^{n_{norm}}_{ij} = logit^n_{ij}/rate^n \label{eq:normal_neg}
\end{equation}

\begin{figure}[t]
  \centering
  \includegraphics[scale=0.4]{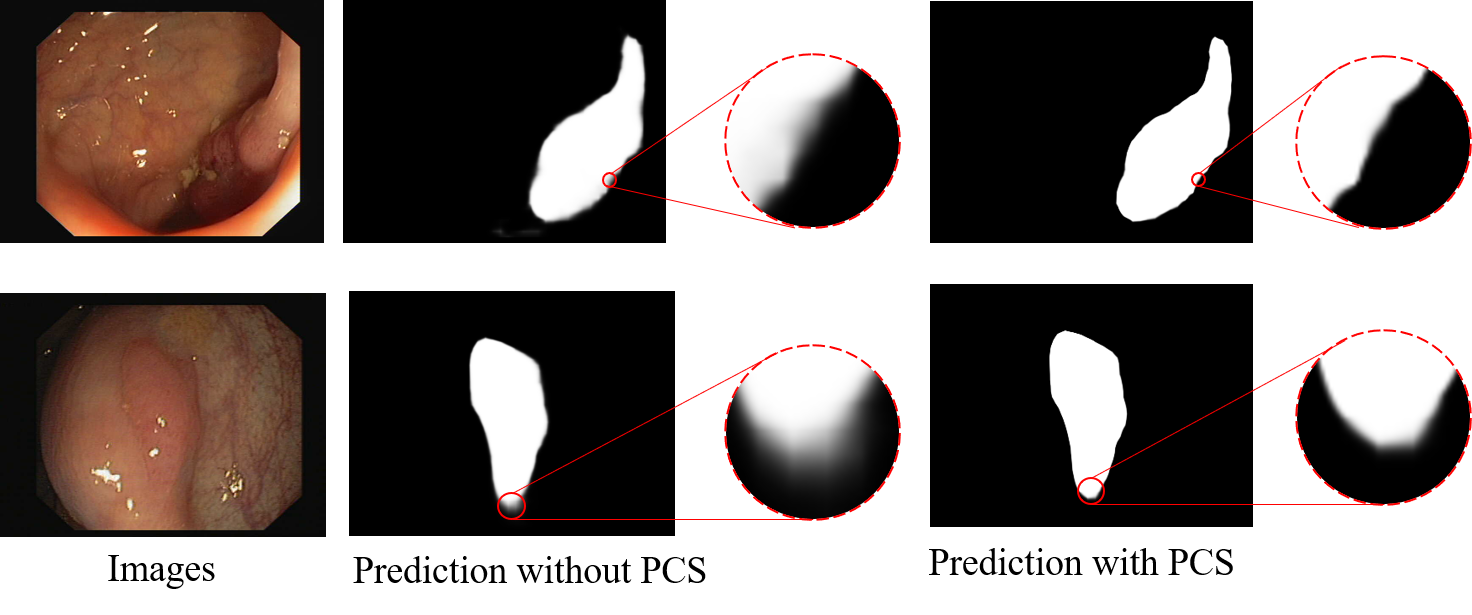}
  \caption{Visual comparison between the predictions with PCS and without PCS.}
  \label{fig:visualization:pcs}
\end{figure}
After normalization, the effect of sample number has been attenuated and the segmentation results will be more accurate. It is worth nothing that PCS is only applied in the inference phase with very little computation cost. Fig.~\ref{fig:visualization:pcs} visualizes some predictions. Obviously, predictions with PCS have clearer boundaries.

\subsection{Loss Function}
We use binary cross entropy (BCE) loss and Dice loss for supervision, as shown in Eq.~\ref{eq:loss}, where $P,G$ are the prediction and ground truth, respectively. $\lambda_1, \lambda_2$ are the weighting coefficients, which are set to 1 for simplification.
\begin{equation}
\label{eq:loss}
Loss = {\lambda_1 BCE(P, G) + \lambda_2 Dice(P, G)}
\end{equation}

\section{Experiments}
\subsection{Datasets and Training Settings}
To evaluate the performance of the proposed SANet, five polyp segmentation datasets are adopted, including Kvasir~\cite{jha2020kvasir}, CVC-ClinicDB~\cite{bernal2015wm}, CVC-ColonDB~\cite{bernal2012towards},  EndoScene~\cite{vazquez2017benchmark} and ETIS~\cite{silva2014toward}. To keep the fairness of the experiments, we follow~\cite{fan2020pranet} advice and take exactly the same training and testing dataset division. Besides, six state-of-the-art methods are used for comparison, namely U-Net~\cite{ronneberger2015u}, U-Net++~\cite{zhou2018unet++}, ResUNet~\cite{zhang2018road}, ResUNet++~\cite{jha2019resunet++}, SFA~\cite{fang2019selective} and PraNet~\cite{fan2020pranet}. We use Pytorch to implement our model. All input images are uniformly resized to 352×352. For data augmentation, we adopt the random flip, random rotation and multi-scale training. The whole network is trained in an end-to-end way, using stochastic gradient descent (SGD). Initial learning rate and batch size are set to 0.04 and 64, respectively. We train the entire model for 128 epoches.

\subsection{Quantitative Comparison}
\begin{table}[t]
\centering
\caption{Performance comparison with different polyp segmentation models. The highest and second highest scores are highlighted in red and blue colors, respectively.}
\label{tab:performace}
\renewcommand\arraystretch{1.2}
\renewcommand\tabcolsep{2pt}
\begin{tabular}{lcccccccccc}
    \hline
        & \multicolumn{2}{c}{Kvasir} & \multicolumn{2}{c}{ClinicDB} & \multicolumn{2}{c}{ColonDB} & \multicolumn{2}{c}{EndoScene} & \multicolumn{2}{c}{ETIS} \\
    \multirow{-2}{*}{Methods} & mDice & mIoU & mDice & mIoU & mDice & mIoU & mDice & mIoU & mDice & mIoU  \\ 
    \hline
    U-Net                       & 0.818 & 0.746 & 0.823 & 0.750 & 0.512 & 0.444 & 0.710 & 0.627 & 0.398 & 0.335 \\
    U-Net++                     & 0.821 & 0.743 & 0.794 & 0.729 & 0.483 & 0.410 & 0.707 & 0.624 & 0.401 & 0.344 \\
    ResUNet                     & 0.791 & -     & 0.779 & -     & -     & -     & -     & -     & -     & -     \\
    ResUNet++                   & 0.813 & 0.793 & 0.796 & 0.796 & -     & -     & -     & -     & -     & -     \\
    SFA                         & 0.723 & 0.611 & 0.700 & 0.607 & 0.469 & 0.347 & 0.467 & 0.329 & 0.297 & 0.217 \\
    PraNet                      & {\color[HTML]{3531FF} 0.898} & {\color[HTML]{3531FF} 0.840} & {\color[HTML]{3531FF} 0.899} &   {\color[HTML]{3531FF} 0.849} & {\color[HTML]{3531FF} 0.712} &  {\color[HTML]{3531FF} 0.640} & {\color[HTML]{3531FF} 0.871} & {\color[HTML]{3531FF} 0.797} & {\color[HTML]{3531FF} 0.628} & {\color[HTML]{3531FF} 0.567} \\
    \hline
    {\color[HTML]{FE0000} SANet(Ours)} & {\color[HTML]{FE0000} 0.904} &  {\color[HTML]{FE0000} 0.847} & {\color[HTML]{FE0000} 0.916} &  {\color[HTML]{FE0000} 0.859} & {\color[HTML]{FE0000} 0.753} & {\color[HTML]{FE0000} 0.670} & {\color[HTML]{FE0000} 0.888} &  {\color[HTML]{FE0000} 0.815} & {\color[HTML]{FE0000} 0.750} & {\color[HTML]{FE0000} 0.654} \\  
    \hline
\end{tabular}
\end{table}

\begin{figure}[t]
  \centering
  \includegraphics[scale=0.42]{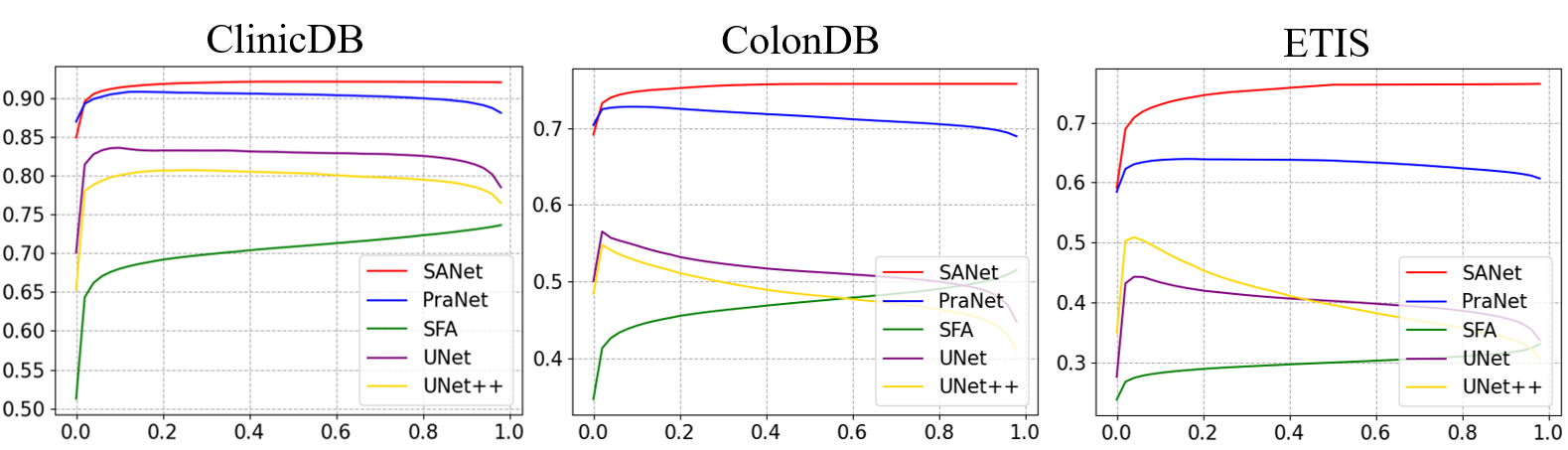}
  \caption{Dice curves under different thresholds on three polyp datasets.}
  \label{fig:dicecurve}
\end{figure}

To prove the effectiveness of the proposed SANet, six state-of-the-art models are used for comparison, as shown in Tab.~\ref{tab:performace}. SANet achieves the best scores across five datasets on both mIoU and mDice, demonstrating the superior performance of the proposed model. In addition, Fig.~\ref{fig:dicecurve} shows the dice values of above models under different thresholds. From these curves, we could observe that SANet consistently outperforms other models, which proves its good capability for polyp segmentation. Furthermore, SANet achieves a speed about 72FPS on RTX 2080Ti GPU, faster than the 64FPS of previous PraNet~\cite{fan2020pranet}.

\subsection{Visual Comparison}
\begin{figure}
  \centering
  \includegraphics[scale=0.42]{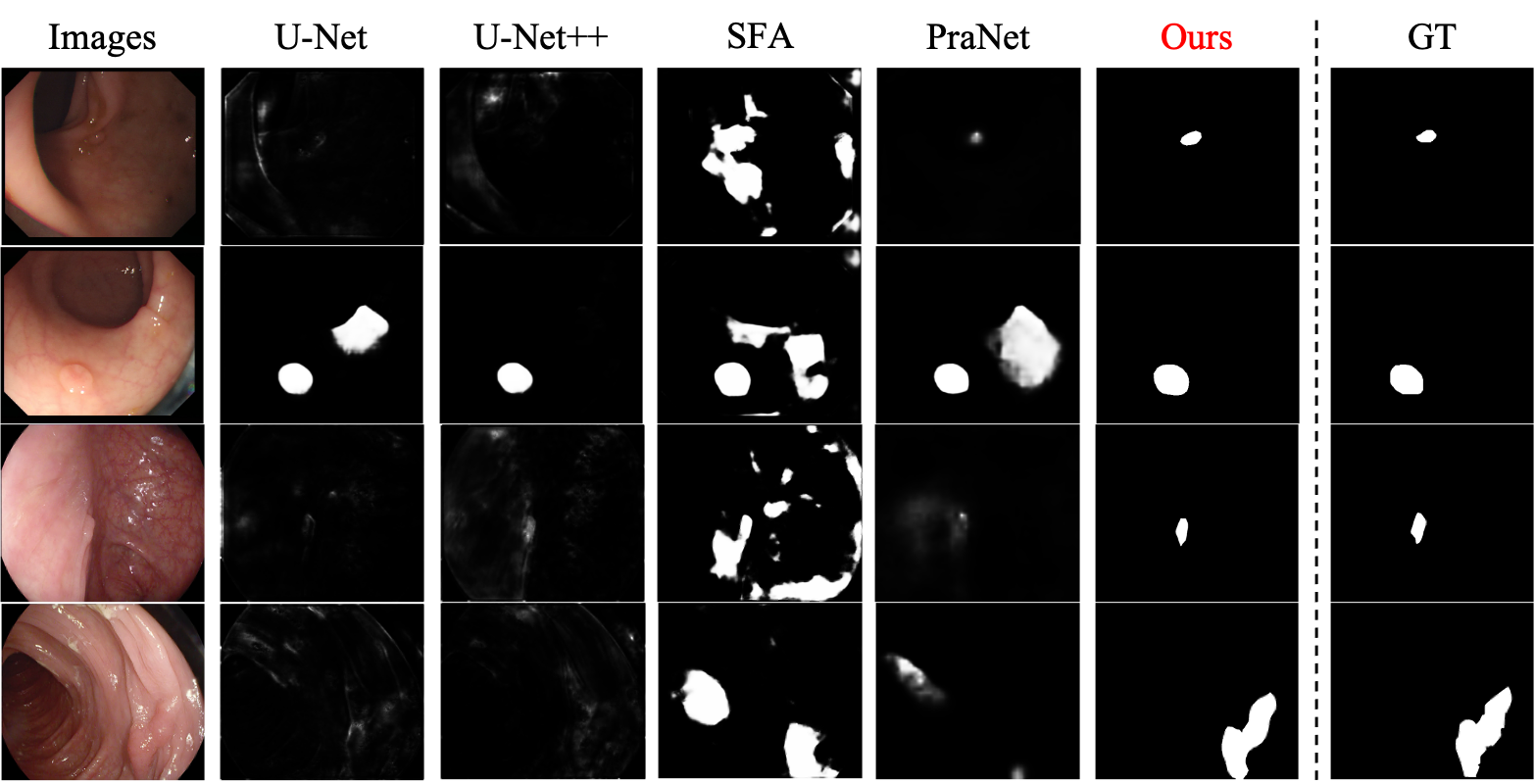}
  \caption{Visual comparison between the proposed method and four state-of-the-art ones.}
  \label{fig:visualization}
\end{figure}

Fig.~\ref{fig:visualization} visualizes some predictions of different models. Compared with other counterparts, our method could not only clearly highlight the polyp regions but also suppress the background noise. Even for the challenging scenarios, our model could handle well and generate accurate segmentation mask.

\subsection{Ablation Study}
To investigate the importance of each component in SANet, both ColonDB and Kvasir datasets are used for controlled experiments. As shown in Tab.~\ref{tab:ablation}, all the modules or strategies are  necessary for the final predictions. Combining all the proposed methods, our model achieves the new state-of-the-art performance.

\begin{table}[t]
  \caption{Ablation study for SANet on the ColonDB and Kvasir datasets.}
  \label{tab:ablation}
  \renewcommand\tabcolsep{10pt}
  \renewcommand\arraystretch{1.1}
  \centering
  \begin{tabular}{l|cc|cc}
    \hline
    \multicolumn{1}{c|}{\multirow{2}{*}{Settings}} & \multicolumn{2}{c|}{ColonDB} & \multicolumn{2}{c}{Kvasir}    \\
    \multicolumn{1}{c|}{}                         & mDice   & mIoU    & mDice   & mIoU    \\
    \hline
    backbone                                      & 0.676   & 0.608   & 0.853   & 0.780 \\
    backbone+SAM                                  & 0.728   & 0.645   & 0.882   & 0.821 \\
    backbone+SAM+CE                               & 0.745   & 0.662   & 0.896   & 0.838 \\
    backbone+SAM+CE+PCS                           & 0.753   & 0.670   & 0.904   & 0.847 \\
    \hline
  \end{tabular}
\end{table}

\section{Conclusion}
Because of the limited dataset, polyp segmentation model is easy to corrupt due to overfitting. In this paper, we attempt to alleviate this problem from two aspects. For the false color causality, we propose to decouple the image color and content by color exchange. For the difficult small polyp segmentation, we design the shallow attention to reduce the data noise. All of these can reduce the interference of irrelevant factors on the model. In the future, we will combine more prior knowledge to design more robust features to remove the interference of independent factors.

\section{Acknowledgement}
The work was supported in part by Key Area R\&D Program of Guangdong Province with grant No.2018B030338001, by the National Key R\&D Program of China with grant No.2018YFB1800800, by Shenzhen Outstanding Talents Training Fund, by Guangdong Research Project No.2017ZT07X152, by NSFC-Youth 61902335, by Guangdong Regional Joint Fund-Key Projects 2019B1515120039, by The National Natural Science Foundation Fund of China (61931024), by helix0n biotechnology company Fund and CCF-Tencent Open Fund.

\bibliographystyle{bibstyle}
\bibliography{bibliography}
\end{document}